\documentclass[american,a4paper,runningheads,envcountsect]{llncs}

\usepackage{babel}
\usepackage[utf8]{inputenc}
\usepackage[T1]{fontenc}

\usepackage{amsmath,amsthm,amssymb}
\usepackage{mathtools}
\usepackage{mathrsfs}
\usepackage{dsfont} 

\usepackage{subcaption}
\usepackage[ruled,vlined,linesnumbered]{algorithm2e}
\usepackage{graphicx}
\usepackage{float} 

\theoremstyle{definition}

\theoremstyle{remark}

\renewcommand{\epsilon}{\varepsilon}
\renewcommand{\phi}{\varphi}

\usepackage[shrink=25,babel,kerning=true]{microtype}
\usepackage{csquotes}
\usepackage{booktabs}
\usepackage{paralist}
\usepackage[subtle]{savetrees}
\usepackage{todonotes}
\presetkeys{todonotes}{color=blue!5}{}

\usepackage[backend=bibtex,style=numeric-comp,doi=false,isbn=false,%
url=false]{biblatex}
\usepackage[colorlinks,citecolor=blue,linkcolor=blue,urlcolor=black,linktocpage]{hyperref}
\usepackage[all]{hypcap}
\usepackage{cleveref}
\let\cref\Cref

\newcommand\blfootnote[1]{%
  \begingroup
  \renewcommand\thefootnote{}\footnote{#1}%
  \addtocounter{footnote}{-1}%
  \endgroup
}

\begin{document}

\title{Formal Context Generation\\ using Dirichlet Distributions
}

\author{Maximilian Felde\inst{1,2} \and Tom Hanika\inst{1,2}}

\date{\today}

\institute{%
  Knowledge \& Data Engineering Group,
  University of Kassel, Germany\\[0.5ex]
  \and
  Interdisciplinary Research Center for Information System Design\\
  University of Kassel, Germany\\[0.5ex]
  \email{felde@cs.uni-kassel.de, tom.hanika@cs.uni-kassel.de}
}
\maketitle

\blfootnote{Authors are given in alphabetical order.
  No priority in authorship is implied.}

\begin{abstract}
  We suggest an improved way to randomly generate formal contexts based on
  Dirichlet distributions.  For this purpose we investigate the predominant way
  to generate formal contexts, a coin-tossing model, recapitulate some of its
  shortcomings and examine its stochastic model. Building up on this we propose
  our Dirichlet model and develop an algorithm employing this idea. By comparing
  our generation model to a coin-tossing model we show that our approach is a
  significant improvement with respect to the variety of contexts
  generated. Finally, we outline a possible application in null model generation
  for formal contexts.
\end{abstract}

\keywords{Formal~Concept~Analysis, Dirichlet~Distribution, Random~Context}

\section{Introduction}
Formal concept analysis (FCA) is often used as a tool to represent and extract knowledge from data sets expressed as cross-tables between a set of objects and a set of attributes, called formal contexts.
Many real-world data sets can be transformed, i.e., scaled, with little effort to be subjected to methods from FCA.
There has been  a great effort to develop methods to efficiently compute both
formal concepts and related properties. This has led to a multitude of algorithms at our disposal.

An important task for the investigation of data sets through FCA is to decide whether observed patterns are meaningful or not.
Related fields, e.g., graph theory and ecology, employ \emph{null model analysis},
cf. \cite{ulrich_gotelli_pattern_detection_in_null_model_analysis}.
This method randomizes data sets with the constraint to preserve certain properties.
Hence, this method may be applied to FCA through randomly generating formal
contexts in a similar manner. One step in this direction is to develop a novel
way to randomly generate formal contexts, which is the topic of this
paper. Randomly generating formal contexts is also relevant for other
applications, e.g., comparing the performance of FCA algorithms, as done in \cite{Kuznetsov_Obiedkov_comparing_algorithms_concept_lattices,bazhanovObiedkov2011comparing}.
However, methods for generating adequate random contexts are insufficiently investigated~\cite{conf/cla/BorchmannH16}.

A naïve approach for a random generation process is to uniformly draw from the
set of all formal contexts for a given set of attributes $M$. This approach is
infeasible as the number of formal contexts with pairwise distinct objects is
$2^{2^{|M|}}$.
A related problem is the random generation of Moore families as investigated in~\cite{conf/cla/Ganter11}.
There, the author suggested an approach to uniformly draw from the set of
closure systems for a given set of attributes. However, this approach is not
feasible on sets of more than seven elements.

The predominant method to randomly generate formal contexts is a coin-tossing
process, mainly due to the ease of use and the lack of proper alternatives.
Yet, this approach is biased to generate a certain class of contexts.
%
Here we step in with a novel approach. 
Based upon a thorough examination of the coin-tossing approach we suggest an improved stochastic model to randomly generate formal contexts using Dirichlet distributions.
For this we analyze the influence of the distribution parameters on the resulting
contexts. Afterwards we empirically evaluate our model on randomly generated
formal contexts with six to ten attributes.
We show that our approach is a significant improvement upon the coin-tossing process in terms of the variety of contexts generated.

As for the structure of this paper in \cref{sec:basics} we first give a short problem description and recall some basic notions from FCA followed by a brief overview of related work in \cref{sec:related_work}.
We proceed by stochastically modeling and examining the coin-toss and suggest the Dirichlet model in \cref{sec:stochastic_modelling}.
In \cref{sec:experiments} we evaluate our model empirically and discuss our findings followed by an outline on the application for null models in \cref{sec:applications}.
Lastly in \cref{sec:conclusion_and_outlook} we give our conclusions and an outlook.

\section{FCA Basics and Problem Description}
\label{sec:basics}
We begin by recalling basic notions from formal concept analysis. For a thorough
introduction we refer the reader to~\cite{GanterWille1999}.  A \emph{formal context} is
a triple $\mathbb{K}\coloneqq(G,M,I)$ of sets.  The elements of $G$ are called \emph{objects} and the
elements of $M$  \emph{attributes} of the context.
The set $I\subseteq G\times M$ is called \emph{incidence relation}, meaning $(g,m)\in I \Leftrightarrow $ the object $g$ has the attribute $m$.
We introduce the common operators, namely the \emph{object derivation}
$\cdot'\colon\mathcal{P}(G)\to\mathcal{P}(M)$ by $A\subseteq G\mapsto A':=
\{m\in M \mid \forall g\in A\colon (g,m)\in I \}$, and the \emph{attribute
  derivation} $\cdot':\mathcal{P}(M)\to\mathcal{P}(G)$ by $B\subseteq M\mapsto
B':=\{g\in G \mid \forall m\in B\colon (g,m)\in I\}$.  A \emph{formal concept} of a
formal context is a pair $(A,B)$ with $A\subseteq G$, $ B \subseteq M$ such
that $A'=B$ and $B'=A$. We then call $A$ the \emph{extent} and $B$ the
\emph{intent} of the concept.  With $\mathfrak{B}(\mathbb{K})$ we denote the set of
all concepts of some context $\mathbb{K}$.  A pseudo-intent of $\mathbb{K}$ is a subset
$P\subseteq M$ where $P\not = P''$ and $Q'' \subseteq P$ holds for every
pseudo-intent $Q\subsetneq P$.
%
In the following we may omit \emph{formal} when referring to formal contexts and formal concepts.
%
Of particular interest in the following is the class of formal contexts called
\emph{contranominal scales}. These contexts are constituted by $([n],[n],\neq)$
where $[n]\coloneqq \{1,\dotsc,n\}$.  The number of concepts for a contranominal
scale with $n$ attributes is $2^{n}$, thus having $2^{n}$ intents and therefore
zero pseudo-intents.  If a context $(G,M,I)$ fulfills the property that for
every $m\in M$ there exists an object $g\in G$ such that $g'=M\setminus\{m\}$,
then $(G,M,I)$ contains a subcontext isomorphic to a contranominal scale of size
$|M|$, i.e, $\exists \hat G\subseteq G$ such that $(\hat G,M,I\cap (\hat G\times
M))\cong ([n],[n],\neq)$ with $n=|M|$.

\begin{figure}[t]
  \centering
  \includegraphics[width=0.5\textwidth]
  {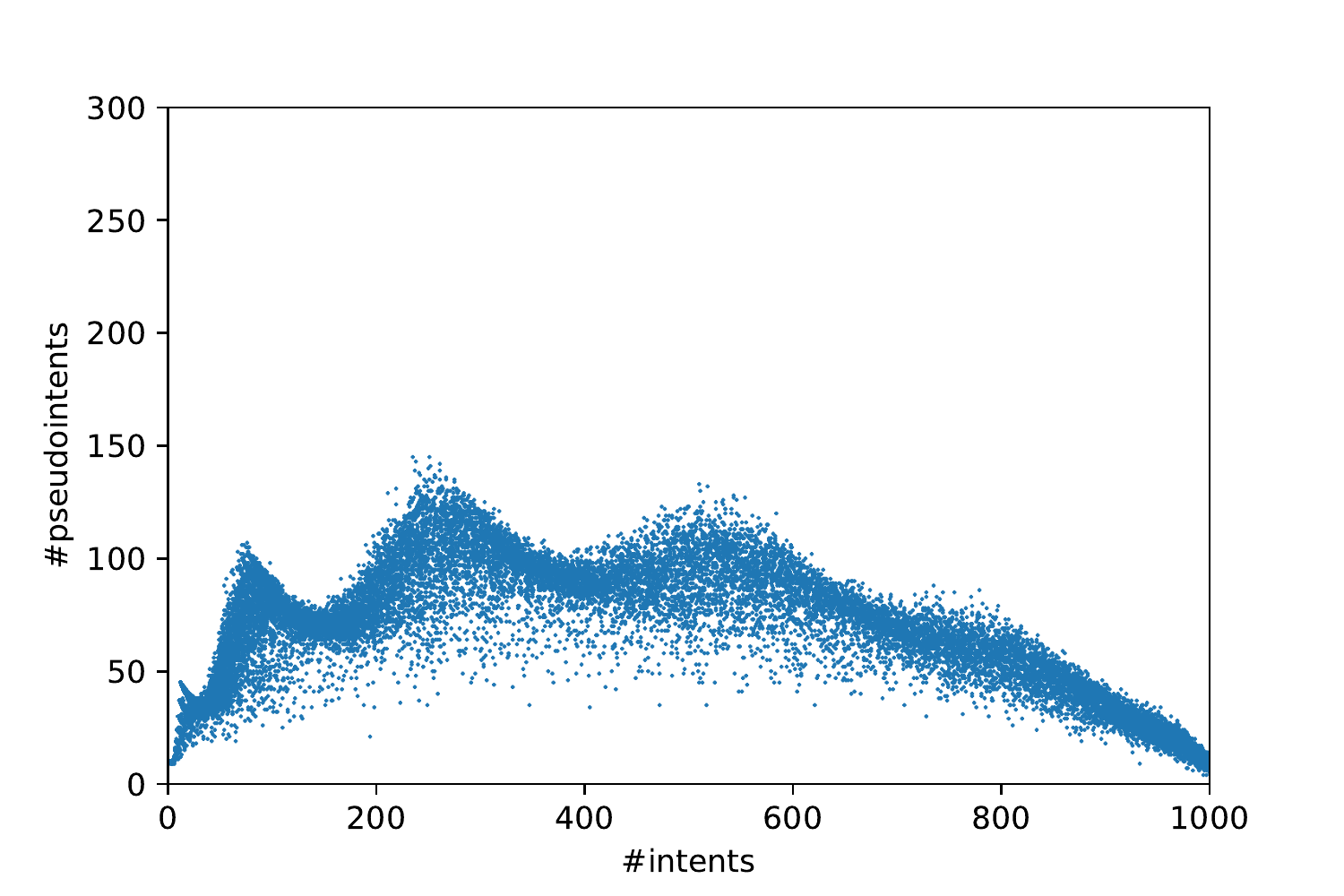}
	\caption{Stegosaurus phenomenon}
	\label{fig:stego0}
\end{figure}

In this paper we deal with the problem of randomly generating a formal context given a set of attributes.
Our motivation originates in~\Cref{fig:stego0} where we show 25000 randomly generated contexts with ten attributes.
The model used to generate these contexts is a \emph{coin-toss}, as recalled
more formally in \cref{sec:stochastic_modelling}. This method is the predominant
approach to randomly generate formal contexts. In~\Cref{fig:stego0} we plotted
the number of intents versus the number of pseudo-intents for each generated
context. We may call this particular plotting method \emph{I-PI plot}, where
every point resembles a particular combination of intent number and
pseudo-intent number, called \emph{I-PI coordinate}. Note that having the same
I-PI coordinate does not imply that the corresponding contexts are
isomorphic. However, different I-PI coordinates imply non-isomorphic formal
contexts.
The reason for employing intents and pseudo-intents is that they correspond to two fundamental
features of formal contexts, namely \emph{concept lattice} and  \emph{canonical
implication base}, which we will not introduce in the realm of this work. 
We observe in~\cref{fig:stego0} that there appears to be a relation between the
number of intents and the number of pseudo-intents. This was first mentioned in
a paper by Borchmann \cite{Borchmann2011}.  Naturally, the question emerges
whether this empirically observed correlation is based on a structural
connection between intents and pseudo-intents rather than chance.  As it turned
out in a later study this apparent correlation is most likely the result of a systematic bias in the
underlying random generation process \cite{conf/cla/BorchmannH16}.

We therefore strive after a novel approach that does not exhibit this or any
other bias.  Consistently with the above the I-PI coordinates and their
distribution are used as an indicator for how diverse created contexts are. The
coin-toss approach will serve as a baseline for this. We start by analyzing the
coin-toss model which leads to a formalization fitted to the requirements of
FCA in~\Cref{sec:stochastic_modelling}. This enables us to discover Dirichlet
distributions as a natural generalization. 


\section{Related Work}
\label{sec:related_work}
The problem depicted in~\Cref{sec:basics} gained not much attention in the
literature so far. The first observation of the correlation between the number
of intents and pseudo-intents in randomly generated contexts was by Borchmann as
a side note in \cite{Borchmann2011}. The phenomenon was further investigated
in~\cite{conf/cla/BorchmannH16} with the conclusion that it is most likely a
result of the random generation process. Their findings suggest that the coin
tossing approach as basis for benchmarking algorithms is not a viable option and
other ways need to be explored. Related to this is a work by Ganter on random
contexts in~\cite{conf/cla/Ganter11}. There the author looked at a method to
generate closure systems in a uniform fashion, using an elegant and conceptually
simple acceptance-rejection model. However, this method is infeasible for
practical use. Furthermore, the authors in~\cite{conf/smc/RimsaSZ13} developed a
synthetic formal context generator that employs a density measure. This
generator is composed of multiple algorithms for different stages of the
generation process, i.e., initialization to reach minimum density, regular
filling and filling close to the maximum density. However, the survey
in~\cite{conf/cla/BorchmannH16} found that the generated contexts exhibited a
different type of systematic bias.

\section{Stochastic Modelling}		
\label{sec:stochastic_modelling}
In the following we analyze and formalize a stochastic model for the coin-toss
approach. By this we unravel crucial points for improving the random generation process.
To enhance the readability we write $Z\sim \mathit{Distribution}$ to denote that
$Z$ is both a random variable following a certain Distribution and a realization
of said random variable.

\subsection{Coin Toss - Direct Model}
Given a set of attributes $M$ we construct $I\subseteq G\times M$ utilizing a
\emph{direct coin-toss} model as follows.
We let $G$ be a set of objects with $|G|\sim \mathrm{DiscreteUniform}[|M|,2^{|M|}]$ and draw a probability $p \sim \mathrm{Uniform}(0,1)$.
For every $(g,m)\in  G\times M$ we flip a binary coin denoted by $\omega_{(g,m)}\sim \mathrm{Bernoulli}(p)$, i.e., $\omega \in \Omega = \{0,1 \}$ where $P(\omega=1)=p$ and $P(\omega=0)=1-p$,
and let
\[
X_{(g,m)}(\omega)\coloneqq
\begin{dcases}
  \{(g,m)\}&\text{if}\ \ \omega = 1\\
  \emptyset&\text{if}\ \ \omega = 0.
\end{dcases}
\]
Then we obtain the incidence relation by $ I\coloneqq\{
X_{(g,m)}(\omega_{(g,m)}) \mid (g,m)\in G\times M\}$. Hence, $I$ contains all
those $(g,m)$ where the coin flip was a success, i.e., $\omega=1$.  If we
partition the set of coin tosses though grouping, i.e., $\{\{ \omega_{(g,m)}
\mid m\in M \} \mid g\in G \}$, and look for some object $g$ at the number of
successful tosses, we see that they follow a $\mathrm{Binomial}(|M|,p)$
distribution. In detail, a binomial distribution with $|M|$ trials and a success
probability of $p$ in each trial.  This means that no matter how $G$, $M$ and
the probability $p$ are chosen, we always end up with a context where the number
of attributes per object is the realization of a $\mathrm{Binomial}(|M|,p)$
distributed random variable for every object $g\in G$.

\begin{figure}[t]
	\includegraphics[width=\textwidth]{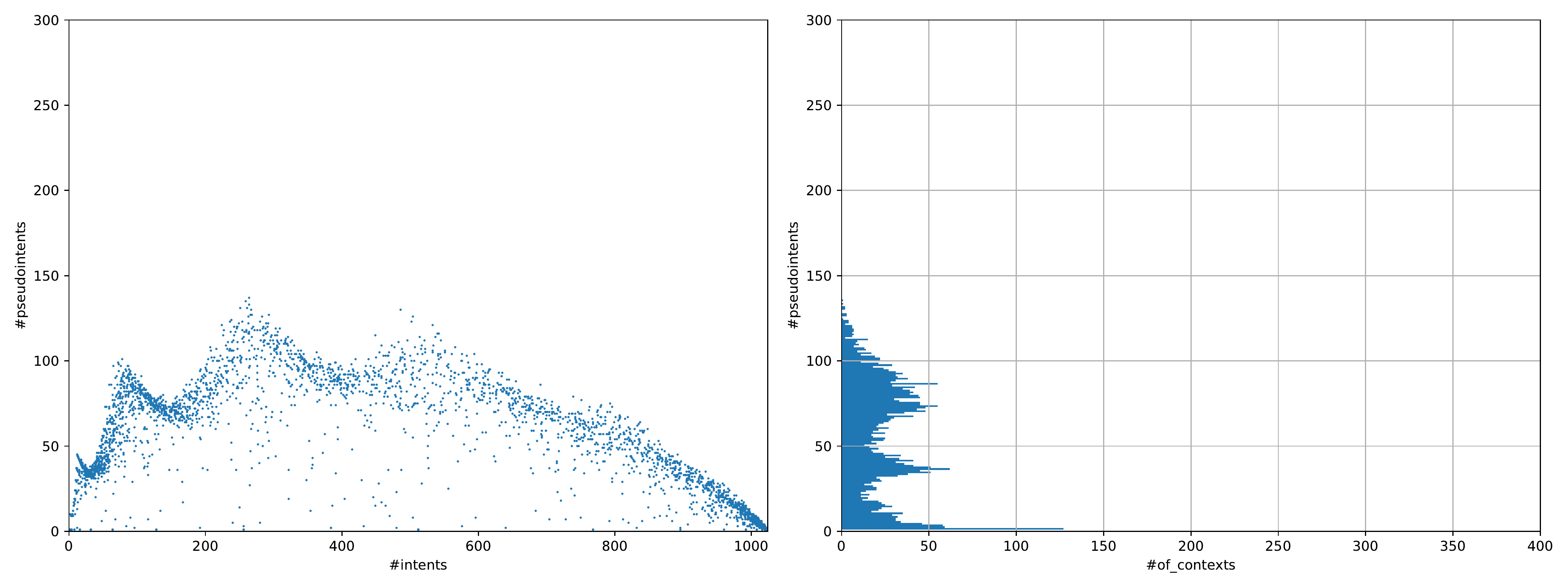}
	\caption{Visualization of I-PI-Coordinates for Coin-Tossing, i.e., Stegosaurus.}
	\label{stego1}
\end{figure}

\begin{example}
\label{ex:coin-toss-stego}
We generated $5000$ contexts with the coin-tossing approach.  A plot of their
I-PI coordinates and a histogram showing the distribution of pseudo-intents are
shown in~\Cref{stego1}. In the histogram we omitted the high value for zero
pseudo-intents. This value emerges from the large amount of generated
contranominal scales by the coin-toss model. In particular, $1714$ of the
contexts contain a contranominal scale and have therefore zero pseudo-intents.

We observe that most of the contexts have less than $100$ pseudo-intents with
varying numbers of intents between $1$ and $1024$.  The majority of contexts has
an I-PI coordinate close to an imaginary curve and the rest has, in most cases,
less pseudo-intents, i.e., their I-PI coordinates lie below this curve.  Looking
at the histogram we observe a varying number of pseudo-intents.  We have a peak
at zero and a high number of $126$ contexts with one pseudo-intent. Additionally
there is a peak of $62$ contexts with $36$ pseudo-intents and a peak of $55$
contexts with $73$ pseudo-intents.  In between we have a low between $18$ to $23$
pseudo-intents and one around $50$.
\end{example}

\subsection{Coin Toss - Indirect Model}
In order to exhibit a critical point in the direct coin-tossing model we
introduce an equivalent model using an indirect approach, called \emph{indirect
  coin-toss}.  Furthermore, this model serves as an intermediate stage to our
proposed generation scheme.

As we just established, the number of successful coin tosses, i.e., number of
attributes per object, follows a binomial distribution.  An indirect model that
generates the same kind of formal contexts as direct coin-tossing can therefore
be obtained by using a binomial distribution.  In contrast to the direct model
we first determine the total number of successful coin-tosses per object and
pick the specific attributes afterwards.
%
We formalize this model as follows.

Given a set of attributes $M$, as before, we let $G$ be a set of objects with $|G|\sim \mathrm{DiscreteUniform}[|M|,2^{|M|}]$ and draw a probability $p \sim \mathrm{Uniform}(0,1)$. 
For every $g\in G$ we let $\theta_g \sim \mathrm{Binomial}(|M|,p)$ be the number
of attributes associated to that object $g$. 
Hence, $\theta_g \in \{0,\ldots,|M|\}$.
We let $\Theta_g\coloneqq\{ B\subseteq M  \mid |B| = \theta_g \}$ be the set of
all possible attribute combinations for $g$ and denote by $\mathrm{DiscreteUniform}(\Theta_g)$ the discrete uniform distribution on $\Theta_g$.
Now for every $g\in G$ we let $B_g \sim \mathrm{DiscreteUniform}(\Theta_g)$ to obtain the set of attributes belonging to the object $g$ and define the incidence relation by
$I\coloneqq\bigcup\{\{(g,m) \mid m \in B_g \}\mid g\in G\}$.
We present pseudocode for the indirect coin-tossing in~\Cref{alg:binomial}.
This serves as a foundation for our proposed generation algorithm in
\cref{subsec:dirichlet_model}.  The indirect formulation reveals that the coin
tossing approach is restricting the class of possible distributions for
$\theta_g$, i.e., the number of attributes for the object $g$, to only the set
of binomial ones.  Thereby it introduces a systematic bias as to which contexts
are being generated.  An example for a context that is unlikely to be created by
the coin-tossing model is a context with ten attributes where every object has
either two or seven attributes.

\begin{algorithm}[t]
	\DontPrintSemicolon
	\SetKwInOut{Input}{Input}
	\SetKwInOut{Output}{Output}
	\Input{a set of attributes $M$}
	\Output{a formal context $(G,M,I)$}
	\BlankLine
	$N\sim \mathrm{DiscreteUniform}[|M|,2^{|M|}]$\;
	$G \coloneqq \{1,\ldots,N\}$\;
	$p\sim \mathrm{Uniform}(0,1)$\;  \label{alg:binom_p}
	
	\ForAll{$g\in G$}
	{
		$\theta_g \sim \mathrm{Binomial}(|M|,p)$\; \label{alg:binom_theta}
		$\Theta_g \coloneqq \{ B\subseteq M \mid |B| = \theta_g \}$\;
		$B_g \sim \mathrm{DiscreteUniform}(\Theta_g)$\;
		$I_g \coloneqq \{ (g,m) \mid m\in B_g\}$\;
	}
	$I\coloneqq \bigcup_{g\in G}I_g$\;
	\Return $(G,M,I)$
	\caption{Indirect Coin-Tossing}
	\label{alg:binomial}
\end{algorithm}

\subsection{Dirichlet Model}
\label{subsec:dirichlet_model}
One way to improve the generating process is to use a broader class of discrete
distributions to determine $\theta_g$ (cf. \Cref{alg:binomial},
\cref{alg:binom_theta}).  In the indirect coin-tossing model we were drawing
from the class of binomial distributions with a fixed number of trials. In
contrast to that we now draw from the class of all discrete probability
distributions on the same sample space, i.e., distributions that have the same
support of $\{0,\ldots, |M|\}$, which in our case represents the possible
numbers of attributes for an object.  For finite sample spaces every probability
distribution can be considered as a categorical distribution.  Therefore, a
common method to draw from the class of all discrete probability distributions
is to employ a Dirichlet distribution. In Bayesian statistics this distribution
is often utilized as prior distribution of parameters of a categorical or
multinomial distribution \cite{ferguson1973}.
One way to define the Dirichlet distribution is to use gamma distributions~\cite{ferguson1973}.
A $\mathrm{Gamma}(\rho,\tau)$ distribution with shape parameter $\rho\geq 0$ and
scale parameter $\tau> 0$ can be characterized on the real line with respect to
the Lebesgue measure by a density function
$$ f(z \mid \rho,\tau) = \frac{1}{\Gamma(\rho)\tau^\rho} \exp^{-z / \tau} z^{\rho-1} \mathds{1}_{(0,\infty)}(z)$$ 
if $\rho>0$, where $\mathds{1}_S$ denotes the indicator function on some set $S$
and $\Gamma$ denotes the gamma function. In the case of $\rho=0$ the gamma
distribution degenerates at zero.
The $\mathrm{Dirichlet}(\beta \boldsymbol{\alpha})$ distribution with parameters $\beta\boldsymbol{\alpha} = (\beta \alpha_1,\ldots,\beta \alpha_K)$, where $\beta > 0$, $\alpha_i\geq0$ for all $i$ and $\alpha_i > 0$ for some $i\in\{1,\ldots,K\}$ and $\sum_{i=1}^{K} \alpha_i =1$, is a probability distribution on the set of $K$-dimensional discrete distributions.
Given independent random variables $Z_1,\ldots,Z_k$ with $Z_i \sim \mathrm{Gamma}(\beta \alpha_i, 1)$ it is defined as the distribution of a random vector $(Y_1,\ldots,Y_K)$ where $Y_i = Z_i / \sum_{j=1}^{k} Z_j$ for $i\in\{1,\ldots,K\}$.
Note that this allows for some of the variables to be degenerate at zero which
will be useful in the application for null models, as we will describe in~\Cref{sec:applications}.
If $\alpha_i > 0$ for all $i$ the random vector $(Y_1,\ldots,Y_K)$ has a density
\begin{equation}
\label{eq:dirichlet_pdf}
f(y_1,\ldots,y_K \mid \beta \alpha_1, \ldots, \beta \alpha_K) = \frac{\Gamma(\beta)}{\prod_{i=1}^{K}\Gamma(\beta\alpha_i)}\prod_{i=1}^{K}y_i^{\beta\alpha_i - 1} \mathds{1}_S(y_1,\ldots,y_K)
\end{equation}
on the simplex $S=\{ (y_1,\ldots,y_K) \in \mathds{R}^{K}  \mid y_i \geq 0, \sum_{i=1}^{K} y_i =1 \}$.
%
%
%
Note that $f$ in \cref{eq:dirichlet_pdf} is a density with respect to the $(K-1)$-dimensional Lebesgue measure and we can rewrite $f$ as a $(K-1)$-dimensional function by letting $y_K = 1 - \sum_{i=1}^{K-1} y_i$ and using an appropriate simplex representation.
Also note that the elements of $(Y_1,\ldots,Y_K)$ have the expected value $\mathds{E}(Y_i)= \alpha_i$, the variance $\mathrm{Var}(Y_i)= \frac{\alpha_i(1-\alpha_i)}{\beta + 1}$ and the co-variance  $\mathrm{Cov}(Y_i, Y_j)= \frac{\alpha_i\alpha_j}{\beta + 1}$ for $i\not= j$.
Hence, the parameter $\boldsymbol\alpha$ is called \emph{base measure} or \emph{mean} as it describes the expected value of the probability distribution and $\beta$ is called \emph{precision parameter} and describes the variance of probability distributions with regard to the base measure. A large value for $\beta$ will cause the drawn distributions to be close to the base measure, a small value will cause them to be distant.
A realization of a Dirichlet distributed random variable is an element of $S$ and can therefore be seen as probability vector of a $K$-dimensional categorical distribution.
\begin{figure}[t]
	\centering
	\begin{subfigure}{.33\textwidth}
		\centering
		\includegraphics[width=\textwidth]{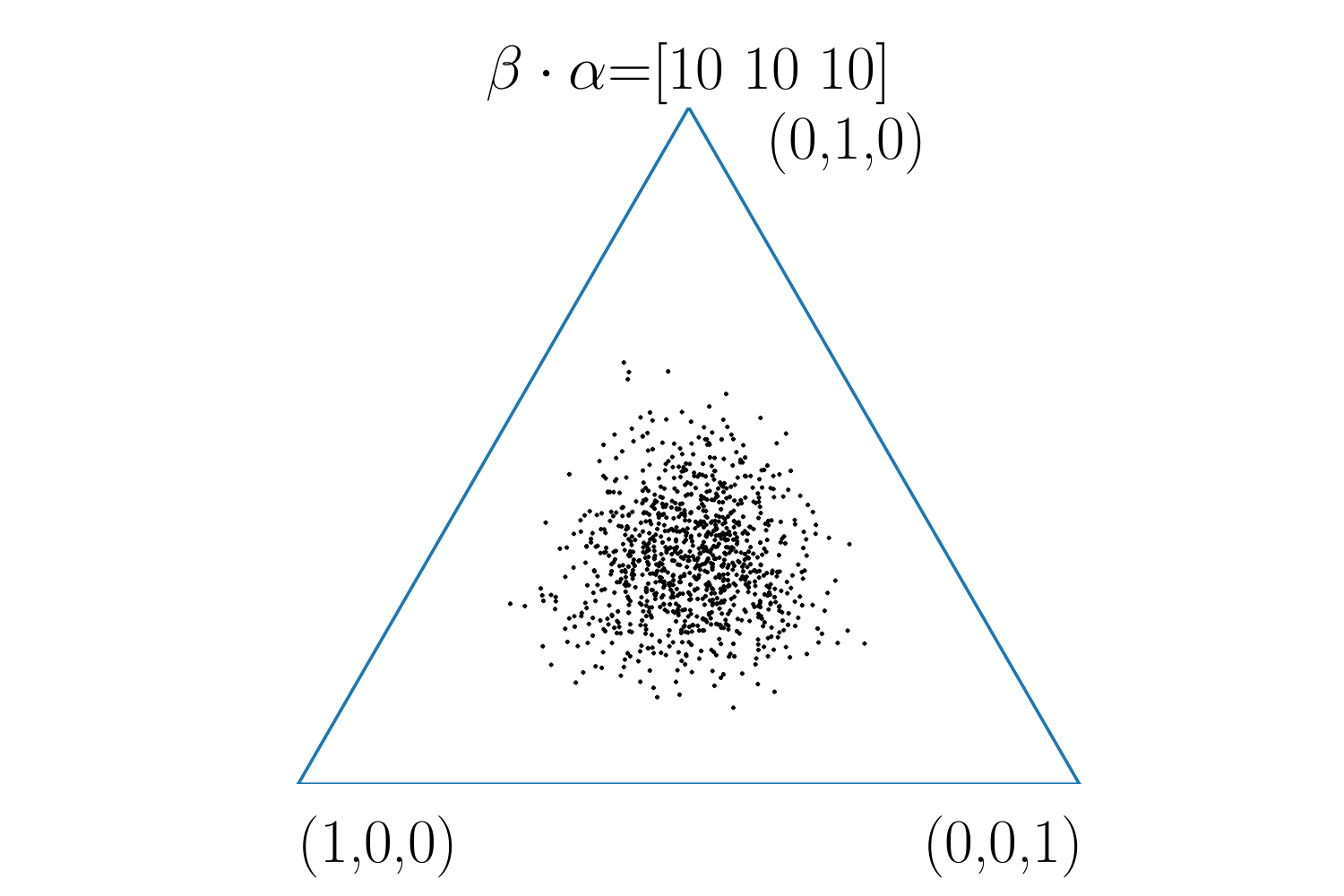}
		\caption{$\beta\cdot\boldsymbol{\alpha} = (10,10,10)$}
		\label{fig:sub1}
	\end{subfigure}%
	\centering
	\begin{subfigure}{.33\textwidth}
		\centering
		\includegraphics[width=\textwidth]{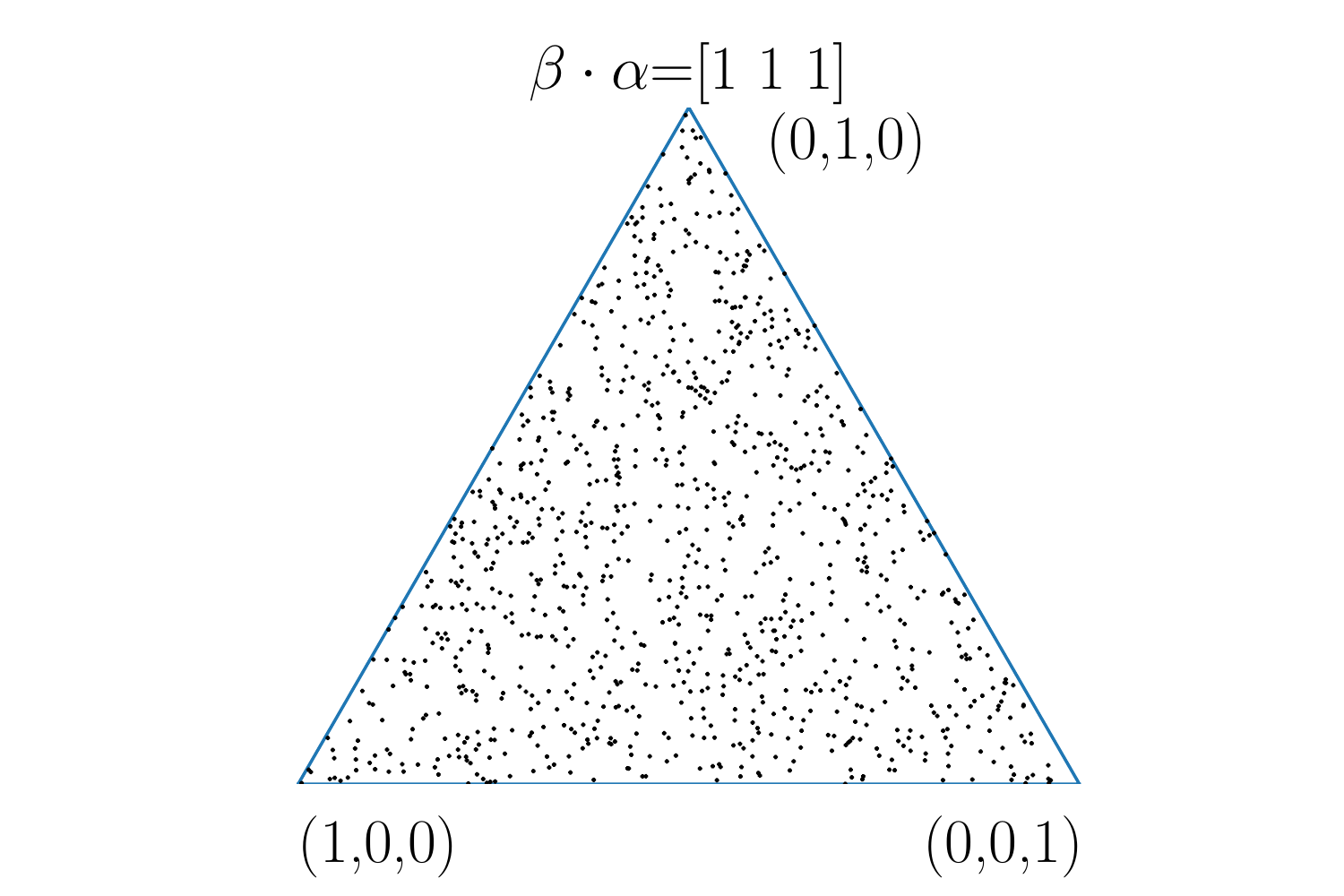}
		\caption{$\beta\cdot\boldsymbol{\alpha} = (1,1,1)$}
		\label{fig:sub2}
	\end{subfigure}%
	\centering
	\begin{subfigure}{.33\textwidth}
		\centering
		\includegraphics[width=\textwidth]{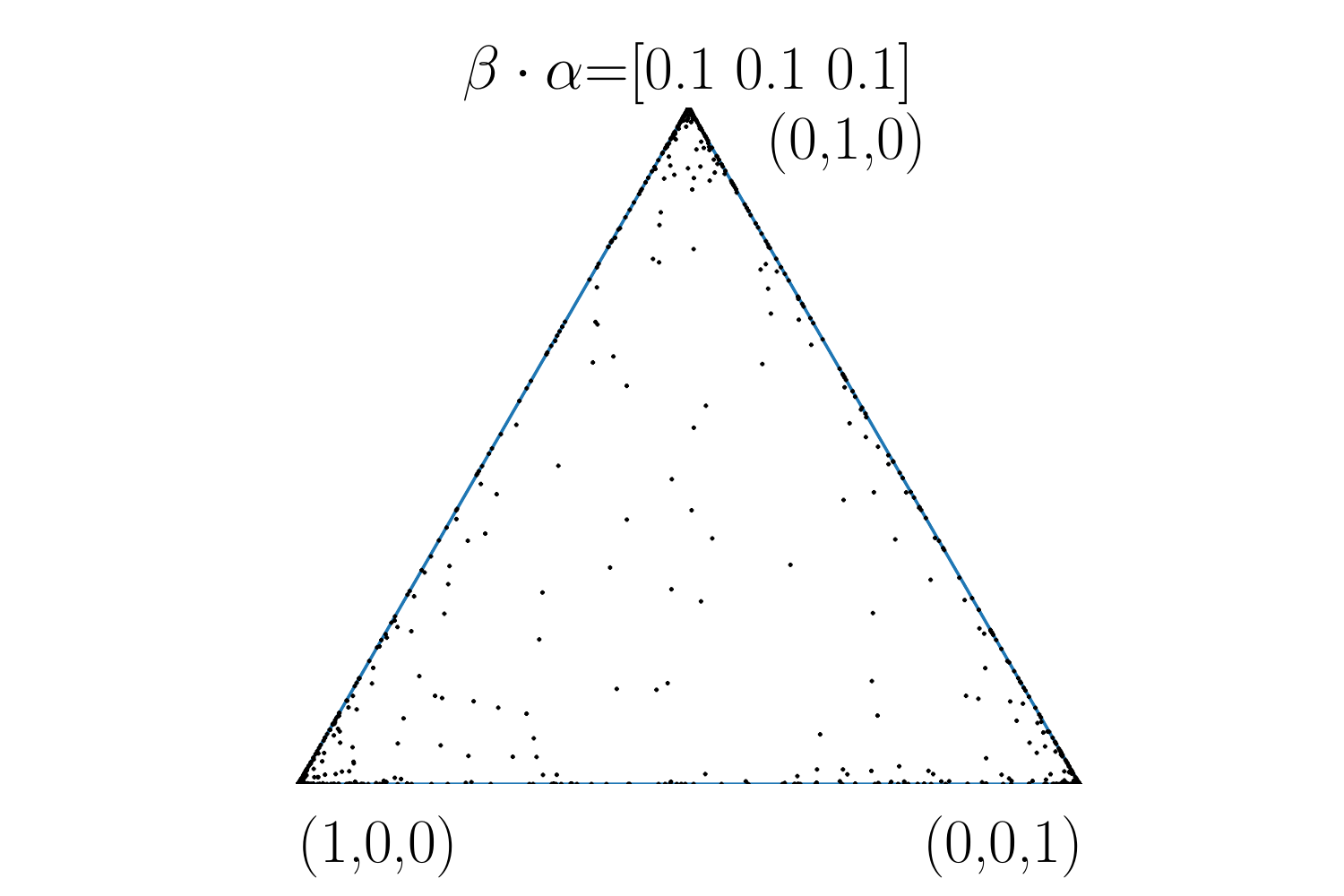}
		\caption{$\beta\cdot\boldsymbol{\alpha} = (0.1,0.1,0.1)$}
		\label{fig:sub3}
	\end{subfigure}%
	\caption{Distribution of categorical probabilities of symmetric Dirichlet distributions.}
	\label{dirichlet-params}
      \end{figure}
      
In~\Cref{dirichlet-params} we illustrate the effects of varying $\beta$. We show
different distributions of probabilities for three categories drawn from
$3$-dimensional Dirichlet distributions.  The base measure $\boldsymbol\alpha$
in each case is the uniform distribution, i.e.,
$(\frac{1}{3},\frac{1}{3},\frac{1}{3})$, the precision parameter $\beta\in\{ 30,
3, \frac{3}{10} \}$ varies.  The choice of $\beta=3$ then results in a uniform
distribution on the probability simplex. For comparison we also chose $\beta=30$
and $\beta=\frac{3}{10}$. A possible interpretation of the introduced simplex is
the following. Each corner of the simplex can be thought of as one category. The
closer a point in the simplex is to a corner the more likely this category is to
be drawn.

In the rest of this section we describe the model for our proposed random formal
context generator.  Given a set of attributes $M$, we let $G$ be a set of
objects with $|G|\sim \mathrm{DiscreteUniform}[|M|,2^{|M|}]$.
%
We then use a probability vector $\boldsymbol p\sim
\mathrm{Dirichlet}(\beta\boldsymbol\alpha)$ to determine the probabilities for
an object to have $0$ to $|M|$ attributes,  where $\boldsymbol\alpha \coloneqq
\nu/||(\nu)||_1$ with $\nu \coloneqq (1,\ldots,1) \in \mathds{R}^{|M|+1}$.
By using $\boldsymbol\alpha$ as base measure and $\beta=|M|+1$, which implies
$\beta\boldsymbol\alpha = (1,\ldots,1)$, we draw uniformly from the set of
discrete probability distributions.
%
As a different way to determine $\theta_g$ we can therefore use $\boldsymbol p=
(p_0,\ldots,p_{|M|})$ as probabilities of a $|M|+1$ dimensional categorical
distribution $\theta_g \sim \mathrm{Categorical}(\boldsymbol p)$. These categories are the numbers of attributes for an object, i.e., $P(\theta_g=c)=p_c$ for $c\in \{0,\ldots,|M|\}$.
Looking back at \Cref{alg:binomial}, \cref{alg:binom_p,alg:binom_theta} we replace the binomial distribution based on a uniformly distributed random variable by a categorical distribution based on a Dirichlet distributed one.
Afterwards we proceed as in \Cref{alg:binomial}.
We present pseudocode for the Dirichlet approach in~\Cref{alg:dirichlet} to emphasize the changes we made compared to \Cref{alg:binomial} and as a further reference for the experiments in \Cref{sec:experiments}. 

\section{Experiments}
\label{sec:experiments}
In this section we present a first experimental investigation of \cref{alg:dirichlet}.
We evaluated the results by examining the numbers of intents and pseudo-intents of generated contexts.
The contexts were generated using \texttt{Python 3} and all further
computations, i.e., the I-PI coordinates, were done using \texttt{conexp-clj}.\footnote{\url{https://github.com/exot/conexp-clj}}
The generator code as well as the generated contexts can be found on GitHub.\footnote{\url{https://github.com/maximilian-felde/formal-context-generator}}

\begin{algorithm}[t]
	\DontPrintSemicolon
	\SetKwInOut{Input}{Input}
	\SetKwInOut{Output}{Output}
	
	\Input{a set of attributes $M$}
	\Output{a formal context $(G,M,I)$}
	
	\BlankLine
	$N\sim \mathrm{DiscreteUniform}[|M|,2^{|M|}]$\; \label{alg:dirichlet_number_objectes}
	$G \coloneqq \{1,\ldots,N\}$\;
	$\nu \coloneqq (1,\ldots,1)$ \tcp*{ ($|M|+1$ ones)}
	$\boldsymbol\alpha \coloneqq \nu / ||\nu||_1$\; \label{alg:dirichlet_alpha}
	$\beta \coloneqq |M|+1$\; \label{alg:dirichlet_beta}
	$\boldsymbol p \sim \mathrm{Dirichlet}(\beta\boldsymbol\alpha)$\; \label{alg:dirichlet_p_from_dir}
	\ForAll{$g\in G$}
	{
		$\theta_g \sim \mathrm{Categorical}(\boldsymbol p)$\;\label{alg:dirichlet_theta_from_cat}
		$\Theta_g \coloneqq \{ B\subseteq M \mid |B| = \theta_g \}$\;
		$B_g \sim \mathrm{DiscreteUniform}(\Theta_g)$\;
		$I_g \coloneqq \{ (g,m) \mid m\in B_g\}$\;
	}
	$I\coloneqq \bigcup_{g\in G}I_g$\;
	\Return $(G,M,I)$
	\caption{Dirichlet Approach}
	\label{alg:dirichlet}
\end{algorithm}

\subsection{Observations}
\begin{figure}[t]
	\includegraphics[width=\textwidth]{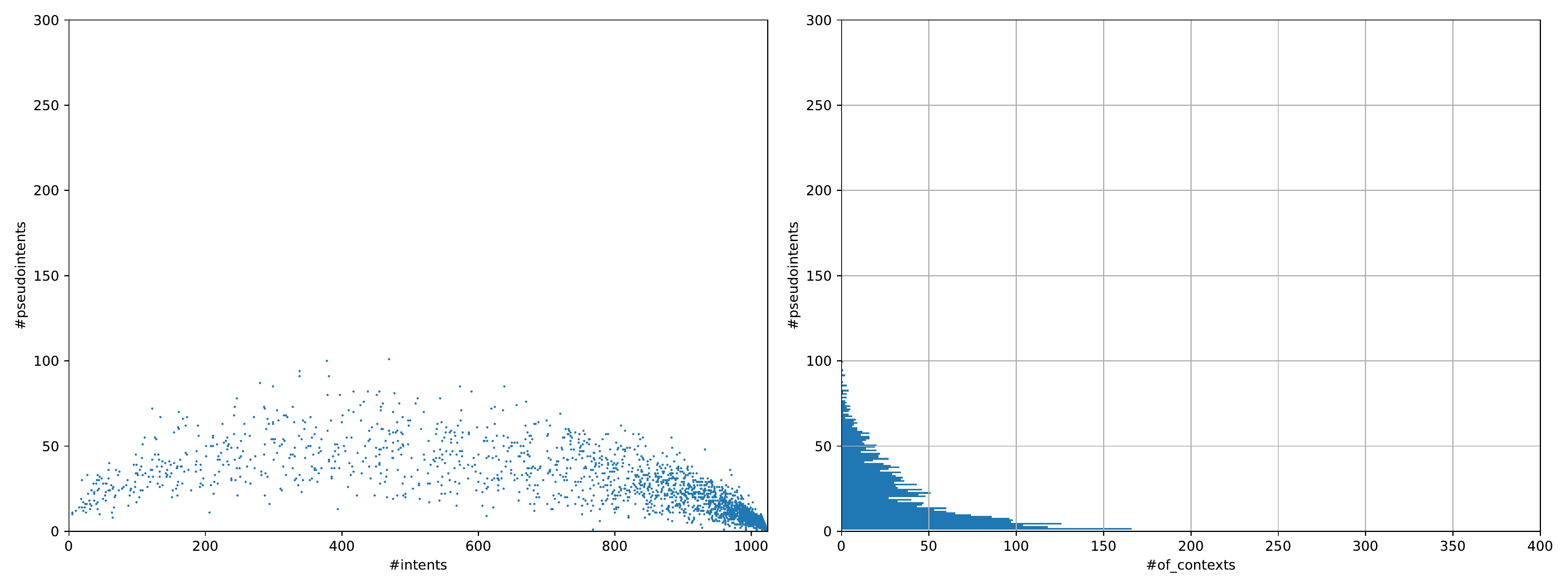}
	\caption{Dirichlet, $\boldsymbol{\alpha} = (\frac{1}{|M|+1},\ldots,\frac{1}{|M|+1})$, $\beta = |M|+1$.}
	\label{dirichlet-stego1}
\end{figure}

For each experiment we generated $5000$ formal contexts with an attribute set $M$
of ten attributes using ~\Cref{alg:dirichlet}. We also employed slightly altered
versions of this algorithm. Those alterations are concerned with the choice of
$\beta$, as we will see in the following. We plotted the resulting I-PI
coordinates and a histogram of the pseudo-intents for each experiment.  In the
histogram we omitted the value for zero pseudo-intents, i.e., the peak for
contexts containing a contranominal scale of size $|M|$.  A comparable experiment
on ten attributes is described in \Cref{ex:coin-toss-stego}, where a (direct)
coin-toss model was utilized. The results of~\Cref{ex:coin-toss-stego} are shown
in~\Cref{stego1}. This will serve as a baseline in terms of variety and
distribution of I-PI coordinates.

First we used \Cref{alg:dirichlet} without alterations.  The results are
depicted in \Cref{dirichlet-stego1}.  We can see that most of the generated
contexts have less than $75$ pseudo-intents and the number of intents varies
between $1$ and $1024$.  There is a tendency towards contexts with fewer
pseudo-intents and we cannot observe any context with more than $101$
pseudo-intents. The number of generated contexts containing contranominal scales
of size $|M|$ was $2438$. The histogram shows that the number of contexts that
have a certain quantity of pseudo-intents decreases as the number of
pseudo-intents increases with no significant dips or peaks.  In this form the
Dirichlet approach does not appear to be an improvement over the coin-tossing
method. In contrary, we observe the spread of the number of pseudo-intents to be
smaller than in~\Cref{ex:coin-toss-stego}.

\begin{figure}[t]
	\includegraphics[width=\textwidth]{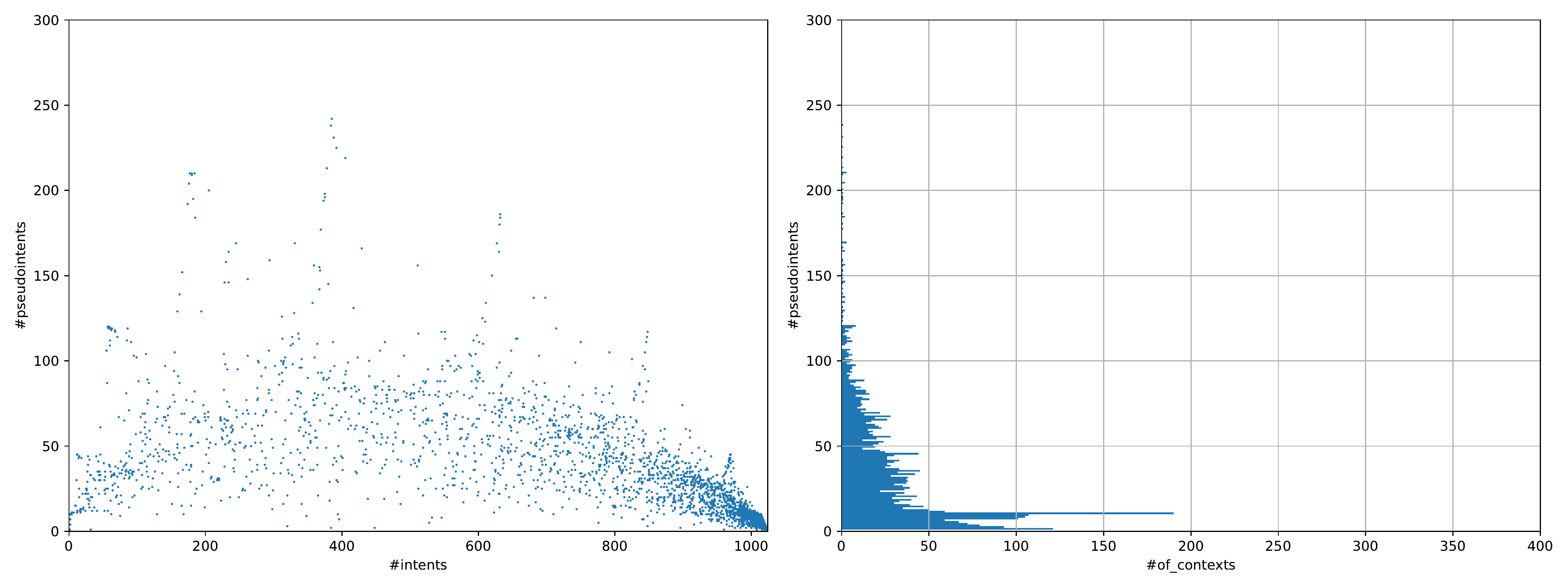}
	\caption{Dirichlet, $\boldsymbol{\alpha} = (\frac{1}{|M|+1},\ldots,\frac{1}{|M|+1})$, $\beta \sim \mathrm{Uniform}(0,|M|+1)$.}
	\label{dirichlet-stego2}
\end{figure}

Our next experiment was randomizing the precision parameter $\beta$ between $0$
and $|M|+1$, i.e., let $\beta \sim \mathrm{Uniform}(0,|M|+1]$ in
\Cref{alg:dirichlet}, \cref{alg:dirichlet_beta}.  We will refer to this
alteration as variation \textbf{A}.  The results are shown in
\Cref{dirichlet-stego2}. We can see that again many contexts have less than
$100$ pseudo-intents and the number of intents once again varies over the full
possible range.  There are $1909$ contexts that contain a contranominal scale of
size $|M|$. However, we notice that there is a not negligible number of contexts
with over $100$ and up to almost $252$ pseudo-intents, which constitutes
theoretical maximum~\cite{conf/cla/BorchmannH16}. Most of these gather around
nearly vertical lines close to $75$, $200$, $380$, $600$ and $820$ intents.
Even though most of the contexts have an I-PI coordinate along one of those
lines there are a few contexts in-between $100$ and $175$ pseudo-intents that do
not fit this description.  Looking at the histogram we can observe again that
while the number of pseudo-intents increases the number of generated contexts to
that pseudo-intent number decreases. This is in contrast
to~\Cref{ex:coin-toss-stego}.  This time, however, we can clearly see a peak at
seven to ten pseudo-intents with $190$ contexts having ten pseudo-intents.
Apart from this we observed no other significant dips or peaks.
%
We also tried randomizing the base measure $\boldsymbol{\alpha}$ using Dirichlet
distributions. However, this did not improve the results.

\begin{figure}[t]
	\includegraphics[width=\textwidth]{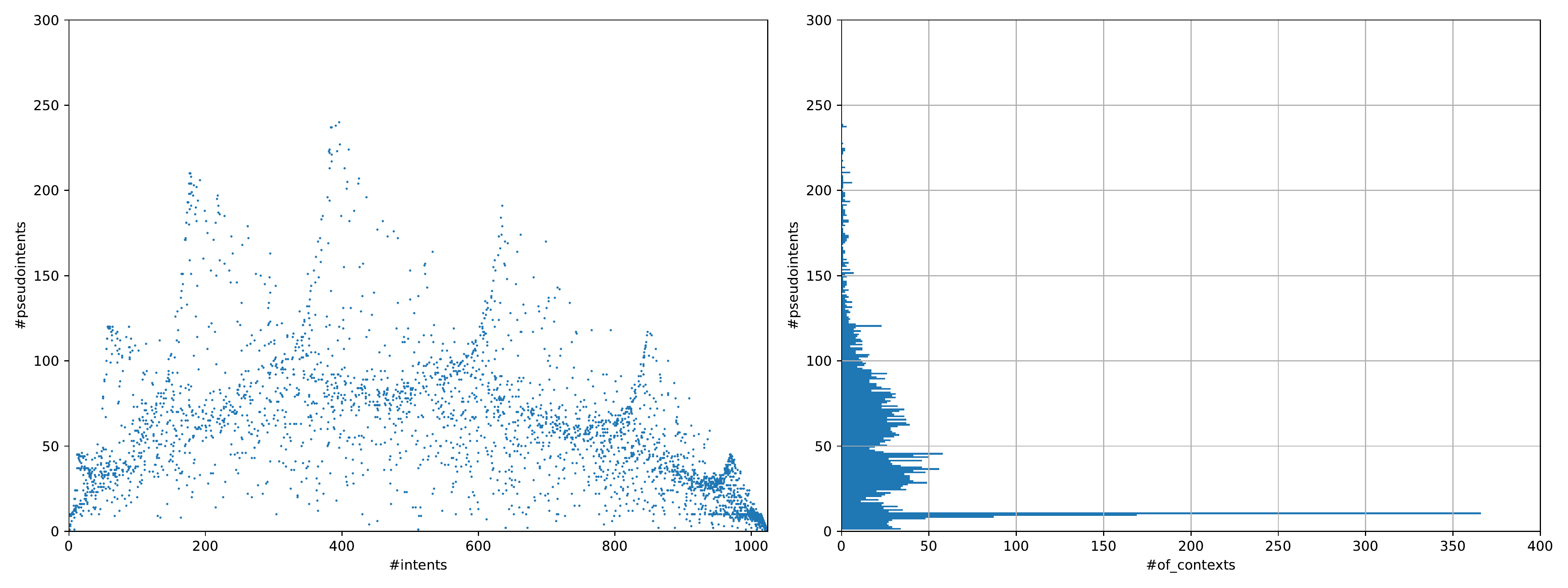}
	\caption{Dirichlet, $\boldsymbol{\alpha} = (\frac{1}{|M|+1},\ldots,\frac{1}{|M|+1})$, $\beta = 0.1(|M|+1)$.}
	\label{dirichlet-stego3}
\end{figure}

Since the last experiment revealed that small values for $\beta$ resulted in a
larger variety of contexts we will now investigate those in more detail. For
this we introduce a constant factor $c$ such that $\beta = c\cdot(|M|+1)$. We
find for the experiment called variation \textbf{B} the factor $c=0.1$ suitable,
as we will explain in~\Cref{sec:discussion}.  A plot of the results can be found
in \Cref{dirichlet-stego3}.  We can see that most of the contexts have less than
$150$ pseudo-intents and the number of intents is between $1$ and $1024$.
Furthermore, the quantity of contexts containing a contranominal scale of size
$|M|$ is $1169$.  This number is about $700$ lower than in variation \textbf{A},
roughly $500$ lower compared to the coin-tossing results
in~\Cref{ex:coin-toss-stego}, and over $1200$ lower than in the unaltered
Dirichlet approach.  We can again observe the same imaginary lines as mentioned
for variation \textbf{A}, with even more contrast. Finally, we observe that the
space between these lines contains significantly more I-PI coordinates. Choosing
even smaller values for $c$ may result in less desirable sets of contexts. In
particular, we found that lower values for $c$ appear to increase the bias
towards the imaginary lines.

The histogram (\Cref{dirichlet-stego3}) of variation \textbf{B} differs
distinguishably to the one in~\Cref{dirichlet-stego2}.  The distribution of
psuedo-intent numbers is more volatile and more evenly distributed. There is a
first peak of $366$ contexts with ten pseudo-intents, followed by a dip to
eleven contexts with seventeen pseudo-intents and more relative peaks of $50$ to
$60$ contexts each at $28$, $36$ and $45$ pseudo-intents. After $62$
pseudo-intents the number of contexts having this amount of pseudo-intents or
more declines with the exception of the peak at 120 pseudo-intents. 

We established that both variations of \Cref{alg:dirichlet} with $\beta \sim
\mathrm{Uniform}(0,|M|+1)$ and $\beta = c\cdot(|M|+1)$ are improvements upon the
coin-tossing approach.  In order to further increase the confidence in our
Dirichlet approach we have generated 100,000 contexts with the coin-tossing
approach as well as with both variations for six, seven and eight attributes. We
compared the distinct I-PI coordinates after certain numbers of generated
contexts. The results of this experiment is shown
in~\Cref{fig:diversity-generated-contexts}.  Each subfigure shows the results
for one attribute set size.  We have plotted the number of generated contexts
versus the number of distinct I-PI coordinates for the coin-toss (green solid
line), variation \textbf{A} (orange dashed line) and variation \textbf{B} (blue
dotted line).  In all three plots we recognize that there is a steep increase of
distinct I-PI coordinates at the beginning followed by a fast decline in new
I-PI coordinates, i.e., a slow increase in the total number of distinct I-PI
coordinates, for all three random generation methods.  The graphs remind of
sublinear growth.  For all three attribute set sizes we can observe that the
graphs of variation \textbf{A} and \textbf{B} lie above the graph of the
coin-toss. Hence, variation \textbf{A} and \textbf{B} generated more distinct
contexts compared to the coin-toss.  Exemplary for $|M|=7$ the coin-tossing
approach resulted in 1963 distinct I-PI coordinates and reached them after
around 99 000 generated contexts. Variation \textbf{A} generated around 19 000
contexts until it hit 1963 distinct I-PI coordinates and reached a total of
around 2450 after 100,000 contexts generated. Variation \textbf{B} reached the
same number of distinct I-PI values already at around 13,000 generated contexts
and resulted in 2550 distinct I-PI coordinates.

\subsection{Discussion}
\label{sec:discussion}
We begin the discussion by relating the parameters of the Dirichlet approach to
the variety of generated contexts.  Afterwards we explore the discrepancy in the
quantities of contexts that contain a contranominal scale and discuss the
observed imaginary lines.  Lastly, we discuss the ability of the different
approaches for generating pairwise distinct I-PI coordinates efficiently.

The Dirichlet approach has two parameters, one being the $\beta$ parameter
related to the variance of the Dirichlet distribution, the other being the
$\boldsymbol{\alpha}$ parameter which describes the expected value of the Dirichlet
distribution, as explained more formally in \cref{subsec:dirichlet_model}.  A
large value for $\beta$ results in categorical distributions that have
probability vectors close to the base measure $\boldsymbol{\alpha}$, following
from the definition. A small value for $\beta$ results in categorical
distributions where the probability vectors are close to the corners or edges of
the simplex, see \Cref{fig:sub3}. As already pointed out
in~\Cref{subsec:dirichlet_model}, those corners of the simplex can be thought of
as the categories, i.e., the possible numbers of attributes an object can have.
This implies that for large $\beta$ the categories are expected to be about as
likely as the corresponding probabilities in the base measure. Whereas, for
small $\beta$ one or few particular categories are expected to be far more
likely than others.

We have seen that the Dirichlet approach without alterations generated around
$2450$ contexts containg a contranominal scale of size $|M|$.  This number was
$1900$ for variation \textbf{A} and $1200$ for variation \textbf{B}.  One reason
for the huge number of contranominal scales generated by the base version of the
Dirichlet approach is that most of the realizations of the Dirichlet
distribution (\cref{alg:dirichlet}, \cref{alg:dirichlet_p_from_dir}) are inner
points of the probability simplex, i.e., they lie near the center of the
simplex.  These points or probability vectors result in almost balanced
categorical distributions (\cref{alg:dirichlet},
\cref{alg:dirichlet_theta_from_cat}), i.e., every category is drawn at least a
few times for a fixed number of draws. This fact may explain the frequent
occurrence of contranominal scales. The expected number of objects with $|M|-1$
attributes that need to be generated for a context to contain a contranominal
scale is low. In more detail, we only need to hit the $|M|$ equally likely
distinct objects, having $|M|-1$ attributes during the generation process.  To
be more precise, the mean $\mu_{N}$ and the standard deviation $\sigma_N$ of the
number of required objects with $|M|-1$ attributes can easily be computed via
$\mu_{N} = N \sum_{k=1}^{N} \frac{1}{k}$ and $\sigma_N^2 = N\sum_{k=1}^{N}
\frac{N-k}{k^2}$ with $N\coloneqq\binom{|M|}{|M|-1}=|M|$,
cf. \cite{Dawkins91/coupon_collector}, as this is an instance of the so-called
\emph{Coupon Collector Problem}.  For example for a context with ten attributes
we get $\mu_{10} \approx 29.3$ and $\sigma_{10} \approx 11.2$, hence we need to
generate on average around $30$ objects with nine attributes to create a
contranominal scale. Although, there is already a high probability of obtaining
a contranominal scale after generating around $18$ objects.  This means if we
generate a context with $|G|=300$ objects and the probability for the category
with nine attributes is around $10\%$ we can expect the context to contain a
contranominal scale. 

If we use a lower value for $\beta$ we tend to get less balanced probability
vectors from the Dirichlet distribution and therefore generate less contexts
that contain a contranominal scale.  The pathological case is a $\beta$ close to
zero, which leads to contexts where all or nearly all objects have the same
number of attributes.  Even then we could expect at least $\frac{1}{|M|+1}$ of
the contexts generated to contain a contranominal scale.  In this case we
basically draw from the set of categories, i.e., from the possible numbers of
attributes. Those are related as corners of the simplex and the probability to
land in the corner belonging to the category of $|M|-1$ attributes is
approximately $\frac{1}{|M|+1}$.

Contexts where every object has the same number of attributes are referred to as
\emph{contexts with fixed row-density} in \cite{conf/cla/BorchmannH16}.  They
were used to show that the coin-tossing approach in practice does not generate a
whole class of contexts.  An explanation for the imaginary lines observed in
\cref{dirichlet-stego2,dirichlet-stego3} is that they correspond to contexts
with fixed row-density, cf. \cite[Fig. 5]{conf/cla/BorchmannH16}.  As pointed
out in the last paragraph very low values of $\beta$ the Dirichlet approach
predominantly generates contexts where all objects belong to few or even only
one category.  This explanation is further supported by the increasing bias of
the context's I-PI coordinates to form those imaginary lines for decreasing
values of $\beta$.  It also accounts for the peak at ten pseudo-intents in the
histograms for variations \textbf{A} and \textbf{B}. This due to the fact that a
fixed row-density context with density $8/10$ that contains all possible objects
has exactly ten pseudo-intents, cf. \cite[Prop. 1]{conf/cla/BorchmannH16}.  This
is again related to the \emph{Coupon Collector Problem}. The solution to this
problem yields the expected number of objects that we need in order to hit every
possible combination. In particular for the case of the peak at ten
pseudo-intents, $N=\binom{10}{8}=45$, $\mu_{45} \approx 198$ and $\sigma_{45}
\approx 56$, meaning if we generate a fixed row-density context with around
$200$ objects we can expect it to contain all possible combinations and
therefore have ten pseudo-intents.  This fits well with the observed $366$
contexts with ten pseudo-intents in variation \textbf{B}.
Consider the case that we only generate fixed row-density contexts containing
all possible attribute combinations and all densities are equally likely.  The
expected number of contexts with eight attributes and therefore ten
pseudo-intents for 5000 generated contexts then is $5000/11 \approx 455$.
Naturally, variation \textbf{B} does not predominantly generate fixed
row-density contexts or even fixed row-density contexts with all possible
attribute combinations. Hence, the afore mentioned $366$ observed contexts with
ten pseudo-intents seem reasonable.

\begin{figure}[t]
	\centering
	\begin{subfigure}{.5\textwidth}
		\centering
		\includegraphics[width=\textwidth]{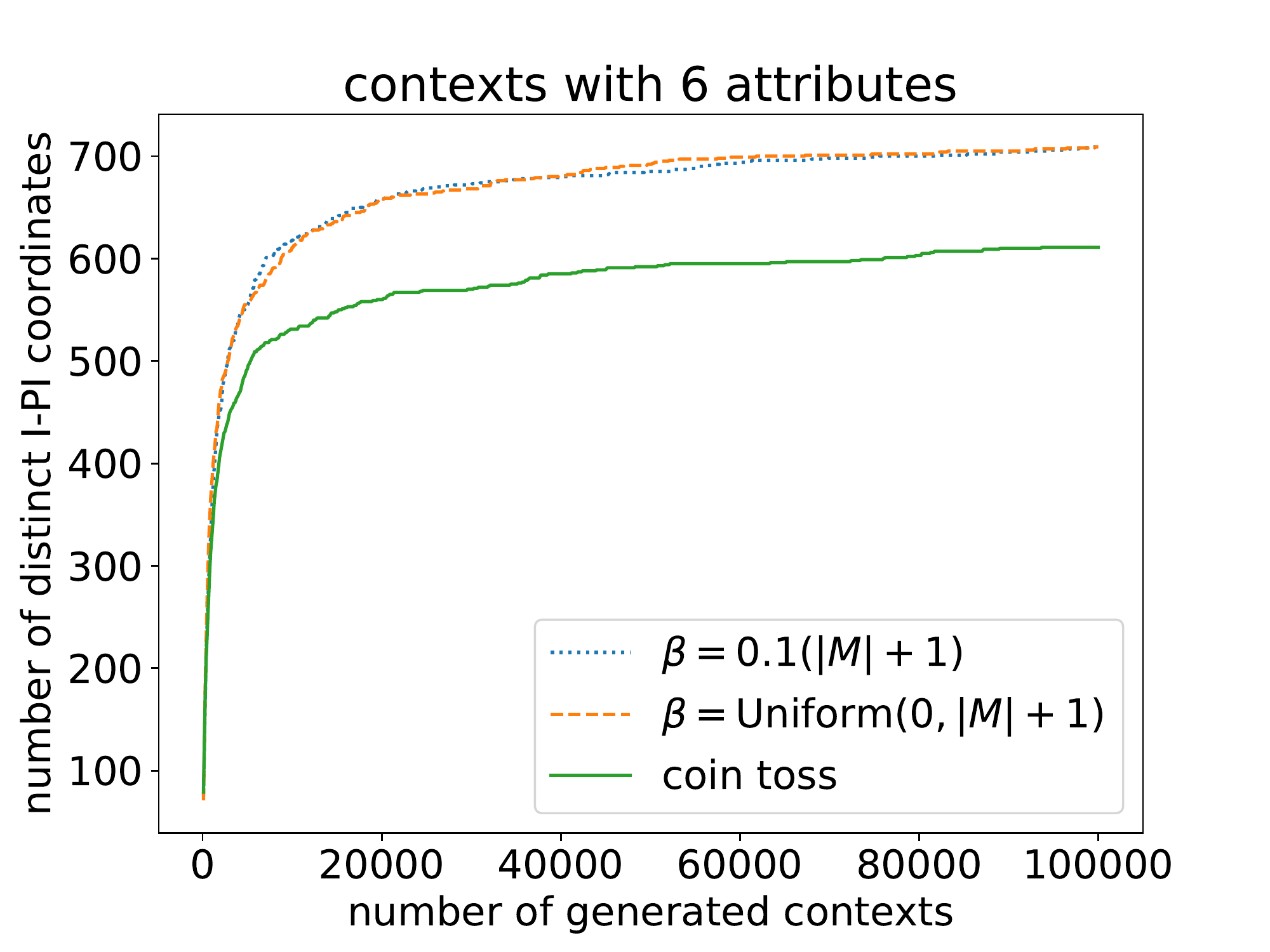}
		\caption{}
		\label{fig:diversity-generated-contexts:sub1}
	\end{subfigure}%
	\centering
	\begin{subfigure}{.5\textwidth}
		\centering
		\includegraphics[width=\textwidth]{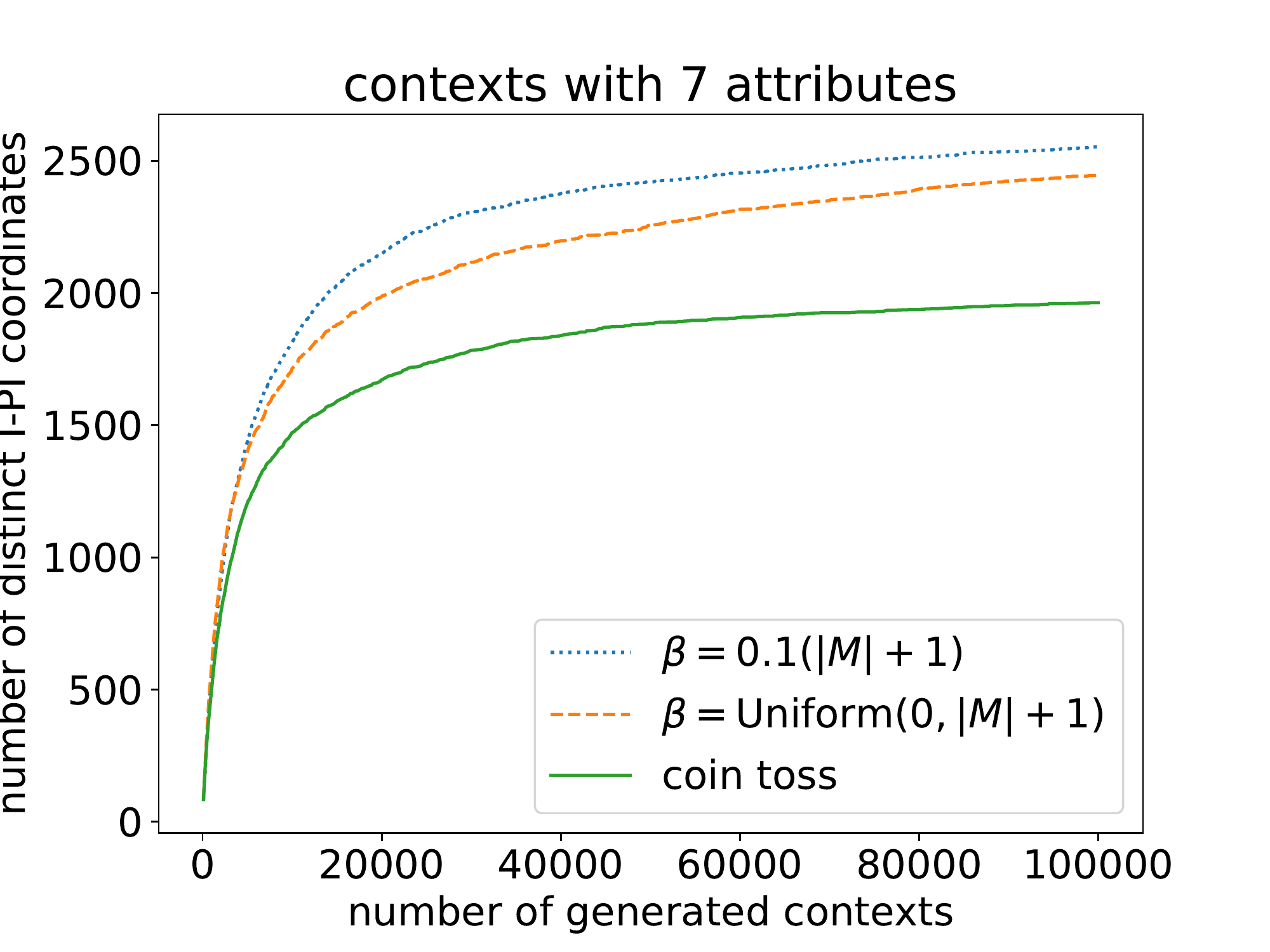}
		\caption{}
		\label{fig:diversity-generated-contexts:sub2}
	\end{subfigure}%

	\centering
	\begin{subfigure}{.5\textwidth}
		\centering
		\includegraphics[width=\textwidth]{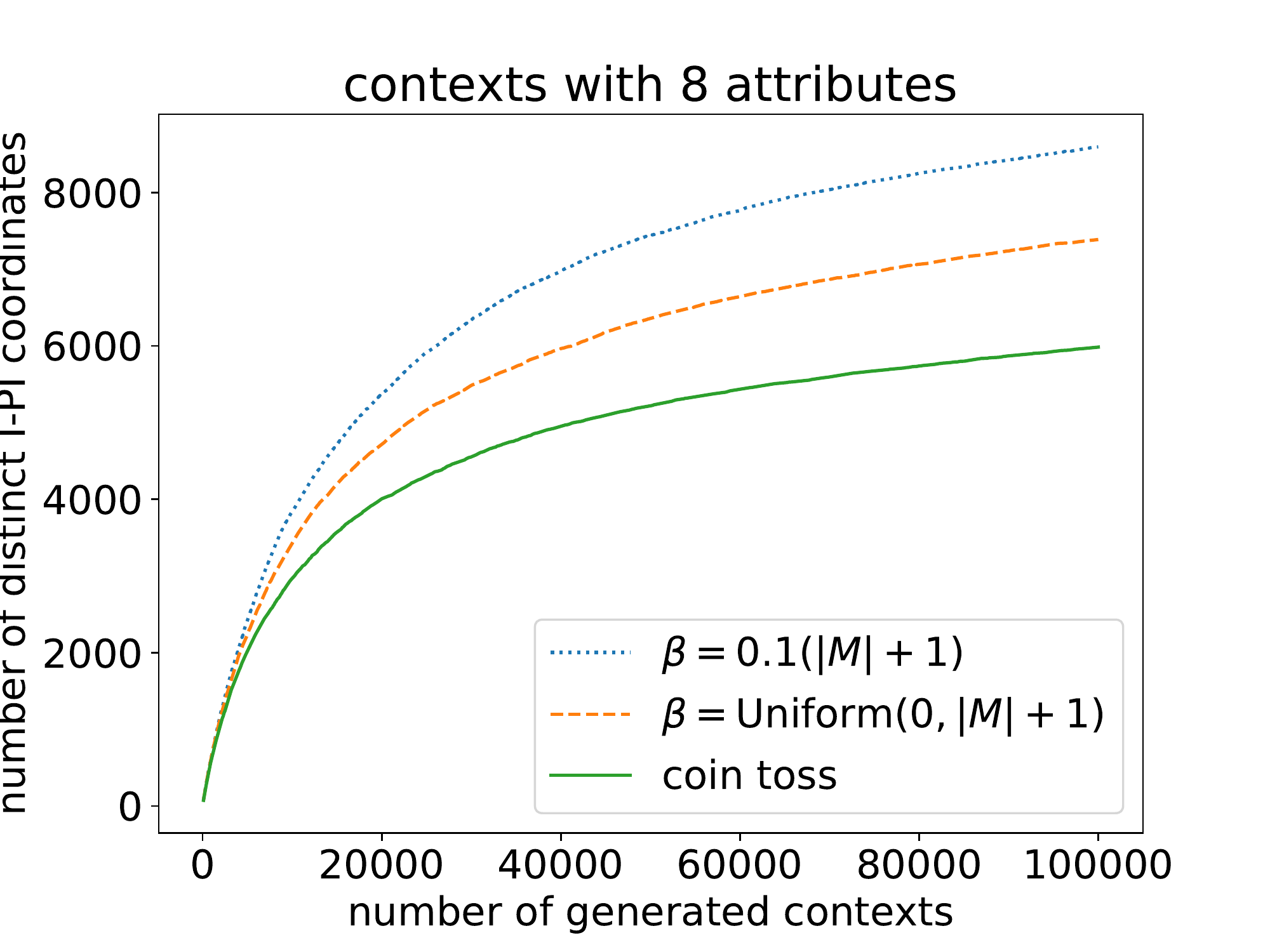}
		\caption{}
		\label{fig:diversity-generated-contexts:sub3}
	\end{subfigure}%
	\caption{Number of distinct I-PI coordinates for up to 100 000 randomly generated contexts with $6$, $7$ and $8$ attributes.}
	\label{fig:diversity-generated-contexts}
\end{figure}

Lastly we discuss the observations from counting distinct I-PI coordinates.
In~\Cref{fig:diversity-generated-contexts} we can see that the Dirichlet
approach results in a broader variety of contexts in comparison to the
coin-tossing for any fixed number of generated contexts.  All three plots show
that there is an initial phase where contexts with new I-PI coordinates are
frequently generated followed by a far longer part where contexts with new I-PI
coordinates become increasingly rare. This is not surprising, since the number
of possible I-PI coordinates is finite and the probability to re-hit increases
with every distinct generated I-PI coordinate. Note that this is an effect that would also be
observed when using a perfectly uniform sampling mechanism. Nonetheless what we
can observe is that it takes the Dirichlet approach significantly longer,
compared to the coin-tossing model, to reach a point where only few new I-PI
coordinates are generated.


\section{Applications}
\label{sec:applications}
We see for our Dirichlet approach at least two major applications.  For one, an
improved random generation process can be used to facilitate more reliable
time performance comparisons of FCA algorithms. The obtained contexts exhibit less
bias and therefore a greater variety concluding in more robust runtime results.
%

A second application is the generation of null models for formal contexts, on
which we elaborate for the rest of this section. Null models are well employed
in graph theory and are an adaption of a statistical concept. The idea there is
to generate random graphs that match some structural properties of some
reference graph.  A null model is therefore not a single graph but denotes a
whole family of graphs that share said properties or an algorithm that generates
such a family.  Given some reference graph, null models are used as a baseline
to decide whether an observed property in the graph is meaningful.  In a
nutshell, the idea of null models is to generate random graphs that are in some
way similar to some reference graph in order to investigate some other property
in this very same reference graph.
There is extensive literature on random graphs available, including a thorough review article by Newman \cite{Newman_structure_and_function_of_complex_networks}.
Two basic structural features in graph theory are the number of nodes and the
degree distribution.  Hence, one way to define a null model for some graph is to
generate random graphs with an equal number of nodes where the degree
distribution as the structural property is preserved. This is a well known idea,
e.g., suggested by Newman when he criticizes the shortcomings of Bernoulli
random graphs.~\cite{Newman_community_structures_in_networks_using_eigenvectors}
An extension to this null model employs expected degree distributions
\cite{Chung2002}. This model is able to capture certain peculiarities arising
from real-world graphs more robustly.

The idea of null models can be translated to the realm of formal contexts in
principle as sketched in the following paragraphs. A possible connection is that
for every formal context one may construct a bipartite graph. The vertex set is
constituted by the disjoint union of the object set and the attribute set. The
edge set is constructed from the incidence relation in a natural way. Applying
now the idea of a graph theoretical null model that preserves the degree
distribution to this bipartite graph can be understood as a family of contexts
where the row- and column-sums are invariant. However, in bipartite graphs t
it is reasonable to only look at one of the elements of the partition. This
relates to analyzing a context with focus on either the attributes or the
objects.  Then a null model can be defined by keeping structural properties,
e.g., preserving the absolute degree distribution, on only one of the elements of
the partition.
%
In the following we focus on row-sums, i.e., the number of attributes per
object. This corresponds very well to our model for randomly generating formal
contexts, cf.~\Cref{subsec:dirichlet_model}.  Of course this choice is
arbitrary to some extend. Interchanging the roles of attributes and objects
would allow to focus on column-sums.

An easy way to generate similar contexts, with respect to row-sum distribution,
is to permute the attributes of every object in the context.  This clearly
leaves the row-sums intact and therefore also the row-sum distribution.
A null model that allows a bit more variation is one where we use the object set
node degree distribution of some reference context. One possibility is to use
that distribution as prior for a categorical distribution considering the node
degrees as categories. From this distribution we then draw for each object in
the reference context a new node degree. Using this we draw accordingly many
elements from the attribute set. Hence, every attribute gets an edge to this
object.  This procedure does not necessarily preserve the object degree
distribution, but the expected distribution equates the distribution of the
reference context.

One further step is to use Dirichlet distributions. We can use the normalized
object degree distribution of the reference context as a base measure for the
Dirichlet distribution. Here we might need the ability to deal with zero
components in the base measure. In this case a large value for $\beta$ is
necessary to control the deviation from the object degree distribution. Again
the realizations of such a Dirichlet distribution are used as priors for a
categorical distribution with the object node degrees as categories. For each
object in the reference context we then generate new node degrees utilizing this
categorical distribution. By definition does the expected object degree
distribution resemble the distribution of the reference context.  For this null
model approach we can employ our results from~\Cref{subsec:dirichlet_model}. One
may apply~\cref{alg:dirichlet} with the appropriate number of objects in
\cref{alg:dirichlet_number_objectes}, the reference contexts object degree
distribution as $\boldsymbol{\alpha}$ in \cref{alg:dirichlet_alpha} and a large
value for $\beta$, e.g., $\beta = 1000\cdot(|M|+1)$, in
\cref{alg:dirichlet_beta}.


%
%

\section{Conclusions and Outlook}
\label{sec:conclusion_and_outlook}

Analyzing a stochastic model for the coin-tossing approach for randomly
generating formal contexts lead in a natural way to a more sophisticated context
generator. By addressing the carved out limitations of the underlying binomial
model we comprehended the versatility of Dirichlet distributions for random
contexts. Based on this we developed an algorithm which can easily be
implemented. This algorithm draws random contexts from a significantly larger
class of contexts compared to the common coin-toss approach.  We empirically
evaluated this new approach for different sizes of attribute sets. The conducted
experiments showed that we generated a significantly broader variety of
contexts. This increased variety may increase the reliability of random context
based investigations, like algorithm performance comparisons.

The novel Dirichlet approach also allows us to generate null models. Basically
in this generation model we can tailor the contexts generated with respect to a
given row-sum-distribution. This method can be employed for empirical
investigations related to formal contexts, like social network analysis with FCA
or ecology.

There are still various unsolved problems remaining. For example, how can one
minimize the amount of generated contranominal scales? This could be done
through considering different base measures. Investigating those 
is itself a fruitful next step in order to understand their relation to
real-world context generation, like null models. Also this
novel approach to random formal contexts raises many more questions about the
relation of the data table context and Dirichlet distributions.


\sloppy
\printbibliography

\end{document}